\pdfoutput=1
\documentclass[11pt]{article}

\usepackage{acl}

\usepackage{times}
\usepackage{latexsym}

\usepackage[T1]{fontenc}
\usepackage[utf8]{inputenc}

\usepackage{microtype}

\usepackage{inconsolata}

\usepackage{graphicx}

\usepackage{amsmath}
\usepackage{amssymb}
\usepackage[skip=0pt]{caption}
\usepackage{subcaption}
\usepackage{pifont}
\usepackage{booktabs}
\usepackage{algorithm}
\usepackage{algpseudocode}
\algrenewcommand\algorithmicrequire{\textbf{Input:}}
\algrenewcommand\algorithmicensure{\textbf{Output:}}
\usepackage{algpseudocode}
\usepackage{threeparttable}
\usepackage{multirow}
\usepackage{tabularx}
\usepackage{cleveref}
\usepackage{enumitem}
\usepackage[textsize=small]{todonotes}
\usepackage{mathtools}
\usepackage{tcolorbox}
\usepackage{soul}
\usepackage{multirow}
\usepackage{colortbl}
\colorlet{lightgray}{White!30!lightgray}
\colorlet{lightblue}{White!70!MidnightBlue}
\newcommand{\cmark}{\ding{51}}
\newcommand{\xmark}{\ding{55}}

\crefname{figure}{Fig.}{Figs.}
\crefname{equation}{Eqn.}{Eqns.}
\crefname{appendix}{Appx.}{Appx.}
\crefname{table}{Table}{Tables}
\crefname{section}{\S}{\S\S}

\newcommand{\interalia}{\emph{inter alia}}
\newcommand{\matrixa}{\textbf{\emph{A}}}
\newcommand{\matrixb}{\textbf{\emph{B}}}
\newcommand{\matrixc}{\textbf{\emph{C}}}

\title{State Space Models are Strong Text Rerankers}

\author{
Jinghua Yan$^{\diamond}$ $^{1}$
Zhichao Xu$^{\diamond}$ $^{1 2}$
Ashim Gupta $^{1}$
Vivek Srikumar $^{1}$\\
$^{1}$ Kahlert School of Computing, University of Utah \quad \\
$^{2}$ Scientific Computing and Imaging Institute, University of Utah \\
{\tt \{jhyan,brutusxu,ashim,svivek\}@cs.utah.edu}
}

\begin{document}
\maketitle

\def\thefootnote{$\diamond$}\footnotetext{Equal Contribution, order decided randomly.}\def\thefootnote{\arabic{footnote}}

\begin{abstract}
Transformers dominate NLP and IR;
but their inference inefficiencies and challenges in extrapolating to longer contexts have sparked interest in alternative model architectures. Among these, state space models (SSMs) like Mamba offer promising advantages, particularly $O(1)$ time complexity in inference.
Despite their potential, SSMs' effectiveness at text reranking\,---\,a task requiring fine-grained query-document interaction and long-context understanding\,---\,remains underexplored.

This study benchmarks SSM-based architectures (specifically, Mamba-1 and Mamba-2) against transformer-based models across various scales, architectures, and pre-training objectives, focusing on performance and efficiency in text reranking tasks. We find that 
(1) Mamba architectures achieve competitive text ranking performance, comparable to transformer-based models of similar size; (2) they are less efficient in training and inference compared to transformers with flash attention; and 
(3) Mamba-2 outperforms Mamba-1 in both performance and efficiency.
These results underscore the potential of state space models as a transformer alternative and highlight areas for improvement in future IR applications.\footnote{The code for reproducing our experiments is available at \url{https://github.com/zhichaoxu-shufe/RankMambaV2}}
\end{abstract}

\section{Introduction}
\label{sec:intro}

The transformer architecture~\cite{vaswani2017attention}  is an established standard within NLP and IR community.
Compared to recurrent neural networks (RNNs)
transformers better capture long-range dependencies and also admit large scale pre-training. 
However, for inference with a sequence of length $L$ and $D$-dimensional hidden states, transformers cost $O(L)$ time and $O(LD)$ space complexity\,---\,proving to be less efficient than RNNs.

Recently, there has been a growing interest in developing alternative architectures for modeling sequence data. For example, RWKV~\cite{peng-etal-2023-rwkv} combines the efficient parallelizable training of transformers with the efficient inference of RNNs. Another notable architecture is the state space model~\cite[SSM, ][]{gu2023mamba,gu2020hippo,gu2021combining}, which is related to convolutional and recurrent neural networks, and also to signal processing literature.

In essence, state space models compress the context into a smaller state of size $N$, achieving $O(1)$ time complexity and $O(ND)$ space complexity in inference time.
However, the capabilities of SSMs are limited by the amount of information that can be compressed in its hidden state. 
To mitigate this, \citet{gu2023mamba} propose a novel selective state space model named Mamba.
Mamba selectively encodes the input to the hidden state to improve model expressiveness, while also addressing the  computation problem with a selective scan method and hardware-aware optimization.
\citet{gu2023mamba} and followup work~\cite[\interalia]{dao2024transformers,zhuvision,wang2024mamba,waleffe2024empirical} examine the efficacy of Mamba models for various sequence modeling tasks, notably language modeling, and also image and audio tasks.
The parameterized SSMs are able to achieve performance close to transformer-based models of similar sizes while also demonstrating efficiency in training and inference.

Despite the growing popularity of state space models, their effectiveness in information retrieval remains underexplored.
Modern search systems typically consist of at least two stages: retrieval and reranking. During retrieval, offline indexes first fetch a preliminary list of candidate documents, which is refined by the reranking model.
Reranking requires  models to understand long context input, and to capture fine-grained query-document interactions. The attention mechanism in the transformer naturally allows for the latter; it allows query tokens to attend to document tokens. 
In contrast, state space models may fail to model long-range dependencies due to their recurrent nature.

In this paper, we examine the following research questions about Mamba-1 and Mamba-2:
\begin{enumerate}[leftmargin=*]
    \item \textbf{Performance RQ}: How do Mamba models compare to transformers for text reranking?    
    \item \textbf{Efficiency RQ}: How efficient are the Mamba architectures with respect to training throughput and inference speed?
\end{enumerate}

To this end, we conduct a rigorous benchmarking study comparing the two  model families, across varying architectures, sizes, pre-training objectives, and attention patterns.
Specifically, we train neural reranking models following established training methodologies outlined in prior literature~\cite{gao2021rethink,boytsov2022understanding,ma2023fine}.
Our experiments allow us to address the two research questions above. 
We find that:
\begin{enumerate}[leftmargin=*]
    \item Mamba-based language models can achieve strong text reranking performance, matching transformer-based models of similar scales. 
    \item Although Mamba architectures have better complexity theoretically, in practice they are less efficient compared to transformer architecture with I/O-aware optimization (e.g., flash attention~\cite{dao2024flashattention}).
    \item Mamba-2 improves upon Mamba-1 in both performance and efficiency.
\end{enumerate}
We discuss the implications of our results and point out future directions of transformer alternative architectures for IR tasks. 

\section{Background: State Space Models}
We will briefly survey state space models and their connection to RNNs and transformers.
We use Structured State Space Sequence Models~\cite[S4,][]{gu2021efficiently} to illustrate the idea behind state space models before describing the Mamba models.

\paragraph{State space models.}
In its simplest form, an SSM maps a 1-dimensional function or sequence $x(t) \in \mathbb{R}$ to $y(t) \in \mathbb{R}$ via a latent state $h(t) \in \mathbb{R}^N$. Here, $t$ denotes a timestep and $N$ is the state size (different from hidden dimensionality $D$). It is parameterized by $(\Delta, \matrixa, \matrixb, \matrixc)$ and defines a \emph{continuous} sequence-to-sequence transformation as:
\begin{align}
    h'(t) &= \matrixa h(t) + \matrixb x(t)  & y(t) &= \matrixc h(t)
\end{align}
The above transformation can be \emph{discretized} as:
\begin{align}
    h_t &= \overline{\matrixa} h_{t-1} + \overline{\matrixb} x_t & y_t &= \matrixc h_t
    \label{eq:rnn}
\end{align}
The discretization of $\overline{\matrixa}$ and $\overline{\matrixb}$ is defined by the \emph{discretization rule}, for example:
\begin{align}
    \overline{\matrixa} &= \exp(\Delta \matrixa) \\ \overline{\matrixb} &= (\Delta \matrixa)^{-1} (\exp(\Delta \matrixa)-I) \cdot \Delta \matrixb
\end{align}
Expanding~\cref{eq:rnn} with the whole sequence $x = (x_1, x_2, \ldots, x_n)$ leads to a convolutional form:
\begin{align}
    y &= x * \overline{\textbf{\emph{K}}}    \label{eq:cnn} \\
    \overline{\textbf{\emph{K}}} &= (\matrixc\overline{\matrixb}, C\overline{\matrixa\matrixb}, \ldots, C\overline{\matrixa}^{n-1} \overline{\matrixb})
\end{align}
While~\cref{eq:rnn} resembles an RNN,~\cref{eq:cnn} looks like a CNN, where $\overline{\textbf{\emph{K}}}$ is a large convolution kernel over the whole input sequence $x$. 
The parameterization of $(\Delta, \matrixa, \matrixb, \matrixc)$ is independent of input sequence $x$ and is fixed during all time steps, a property referred to as linear time invariance (LTI).
Structured state space models (S4) imposes a structure on the $\matrixa$ matrix for efficiency.
Existing works~\cite{gu2021efficiently,gupta2022diagonal,smith2022simplified}  employ a diagonal matrix, thus $\matrixa \in \mathbb{R}^{N \times N}$, $\matrixb \in \mathbb{R}^{N \times 1}$ and $\matrixc \in \mathbb{R}^{1 \times N}$ matrices are all represented by $N$ parameters. 

The above expressions can be generalized to $D$-channel features, i.e., $x_t, y_t \in \mathbb{R}^D$. A concrete example might be $D$-dimensional word embeddings or hidden states. 
In this case, computation of $\matrixa, \matrixb, \matrixc$ is applied to each channel independently. 

\paragraph{Mamba-1 Models.}
State space models compress potentially unbounded context into a state $h_t \in \mathbb{R}^N$, potentially limiting their effectiveness.
\citet{gu2023mamba} propose to make the parameters $(\Delta, \matrixb, \matrixc)$ input-dependent. 
This modification changes the model from time-invariant to time-varying, therefore posing challenges to the model's computational efficiency; the model now cannot be trained in CNN mode.
\citet{gu2023mamba} address this via a hardware-aware optimization algorithm called \emph{Selective Scan}. We refer the reader to the original paper for details.

\paragraph{Scalar Structured SSM.} 
Mamba-2~\cite{dao2024transformers} restricts the matrix $\matrixa$ to be
a scalar times identity matrix; i.e., all the diagonal elements of $\matrixa$ are the same value.
It also introduces a new hyperparameter $P$, the SSM head dimension, which is analogous to the transformer head dimension, i.e., $D = P \times \texttt{\#heads}$. Mamba-2 uses different $(\Delta,\textbf{\textit{A}},\textbf{\textit{B},\textbf{\textit{C}}})$ for each SSM head, and $P$ is set to 64 or 128, similar to transformers. Further, \citet{dao2024transformers} develop efficient implementations for training and inference, enabling much larger state size (from $N=16$ in Mamba-1 to $N=64, 256$ or larger in Mamba-2), while simultaneously being faster in training. Subsequent works~\cite[\interalia]{yang2024parallelizing,qin2024hgrn2,dao2024transformers} also reported Mamba-2's performance and efficiency improvement over Mamba-1. 

\section{The Text Reranking Problem}
Modern IR systems employ a two-stage retrieval-reranking pipeline~\cite[\interalia]{schutze2008introduction,zhang-etal-2021-learning-rank,asai2024openscholar,xu2025survey}. 
After the initial retrieval by an efficient, scalable first-stage retriever, a reranker refines the ranklist to optimize ranking metrics.
Reranking involves ordering texts (passages or documents) by their \emph{relevance} to a query, with passage reranking being a finer-grained form of document reranking. 
Our focus is to study this second stage and perform a comprehensive analysis of different rerankers for the tasks of both passage reranking and document reranking. 

Let $q$ be an input query, and $d \in \mathcal{D}$ be a text, where $\mathcal{D}$ is the set of all texts (passages for passage reranking and documents for document reranking).
The reranking model $f_{\theta}(q,d)$, parameterized by $\theta$, predicts a scalar relevance score.
The model $f$ is instantiated as a linear layer on top of a language model.
We adopt the common practice of concatenating the query and the document as input to the model~\cite[\interalia]{nogueira2019document,yates2021pretrained,boytsov2022understanding,ma2023fine}.

\paragraph{Training a Reranker.}
Training the reranking model involves sampling negatives from the document collection. We use the recommended setup from literature~\cite{gao2021rethink,ma2023fine,boytsov2022understanding,xu2024rankmamba} to sample hard negatives from the retrieval results obtained from the first-stage retriever.

Let us denote the relevant document to query $q_i$ as $d_i^{+}$, and sampled negatives as $d_i^{-} \in \mathcal{D}_i^{-}$, training pair $(q_i, d_i^{+}) \in \mathcal{S}$, the reranking model is trained with optimizing the following softmax loss:
\begin{equation}
    \resizebox{0.465\textwidth}{!}{$-\frac{1}{|\mathcal{S}|} \sum\limits_{(q_i, d_i^{+})\in \mathcal{S}} \log  \frac{\exp f_{\theta}(q_i, d_i^{+})}{\exp f_{\theta}(q_i, d_i^{+}) + \sum_{j\in \mathcal{D}_i^{-}} \exp f_{\theta}(q_i, d_i^{-})}$}
\end{equation}
We pack multiple training instances into a minibatch and jointly optimize the backbone language model and the linear layer.
\section{Experiments}
In this section, we describe the experimental setup for passage  (\cref{subsec:passage}) and document reranking (\cref{subsec:doc_ranking}). 
For \textbf{Performance RQ}, we report results and analyze the implications in~\cref{subsec:passage_results} and~\cref{subsec:doc_results} respectively. 
Then we address \textbf{Efficiency RQ} in~\cref{subsec:efficiency_results}.

\subsection{Passage Reranking}
\label{subsec:passage}
First, let us examine the passage reranking task.

\paragraph{Datasets and Evaluation Metrics.} We employ the passage ranking subset of the well-known MS MARCO dataset~\citep{bajaj2016ms} which contains a total of 524K training instances. For the passage retriever in the first stage, we use \texttt{BGE-large-en-v1.5}~\cite{xiao2023bge} due to its strong trade-off between retrieval performance and size of the retriever. (See~\cref{tab:results_passage_retrieval} in the appendix for this comparison). 
% \footnote{\url{https://huggingface.co/BAAI/bge-large-en-v1.5}}
The training set for our passage reranker is constructed by uniformly sampling 15 hard negatives from the ranklist of top-1000 passages returned by the BGE retriever.~\citet{zhuang2023rankt5,ma2023fine} highlight that increasing the number of negatives along with the global batch size leads to better ranking performance. Training demands significant GPU RAM. We determined these hyperparameters by balancing performance and the available hardware resources.

\begin{table}[t]
% \vspace{0pt}
\centering
\resizebox{\columnwidth}{!}{
\begin{tabular}{
p{0.35\columnwidth}
p{0.2\columnwidth}
p{0.3\columnwidth}
p{0.15\columnwidth}
}
\toprule
\begin{tabular}[c]{@{}l@{}l} \textbf{Model} \\ \, \\ \end{tabular} &
\begin{tabular}[c]{@{}l@{}l} \textbf{Size} \\ \, \\ \end{tabular} & 
\begin{tabular}[c]{@{}l@{}l} \textbf{Architecture} \\ \, \end{tabular} & 
\begin{tabular}[c]{@{}l@{}l} \textbf{Pre-train} \\ \textbf{\#Tokens} \\ \end{tabular} \\
\rowcolor{lightgray}
\multicolumn{4}{l}{\emph{\textbf{Encoder-only Models} (Bi-directional)}} \\
BERT-base & 110M & Transformer & 3.3B \\
RoBERTa-base & 120M & Transformer & 33B \\
ELECTRA-base & 105M & Transformer & 3.3B \\ \hline
BERT-large & 330M & Transformer & 3.3B \\
RoBERTa-large & 335M & Transformer & 33B \\
ELECTRA-large & 320M & Transformer & 33B \\

\rowcolor{lightgray}
\multicolumn{4}{l}{\emph{\textbf{Encoder-Decoder Models} (Bi-directional)}} \\
BART-base & 130M & Transformer & 33B \\ \hline
BART-large & 385M & Transformer & 33B \\

\rowcolor{lightgray}
\multicolumn{4}{l}{\emph{\textbf{Decoder-only Models} (Uni-directional)}} \\
OPT-125M & 125M & Transformer & 180B \\
Mamba-1-130M & 130M & Mamba-1 & 300B \\
Mamba-2-130M & 130M & Mamba-2 & 300B \\ \hline 
OPT-350M & 350M & Transformer & 180B \\
Mamba-1-370M & 370M & Mamba-1 & 300B \\
Mamba-2-370M & 370M & Mamba-2 & 300B \\ \hline 
Mamba-1-790M & 790M & Mamba-1 & 300B \\
Mamba-2-780M & 780M & Mamba-2 & 300B \\ \hline 
OPT-1.3B & 1.3B & Transformer & 180B \\
Mamba-1-1.4B & 1.4B & Mamba-1 & 300B \\
Mamba-2-1.3B & 1.3B & Mamba-2 & 300B \\
Llama-3.2-1B & 1.3B & Transformer++ & 15T \\

\bottomrule
\end{tabular}
}
\caption{Details of the Pre-trained language models used in our comparative study. Transformer++ indicates the state-of-the-art transformer architecture. Note the pre-training \#tokens is not directly comparable between encoder-only, encoder-decoder and decoder-only models due to different pre-training objectives.
}
\label{tab:models}
% \vspace{-5pt}
\end{table}

The in-domain evaluation is conducted using the official passage ranking development set (\texttt{Dev}) containing 6,980 queries. We also include TREC DL19/DL20~\cite{craswell2020overview,craswell2021overview} evaluation set that contains 43/54 queries with in-depth annotation for passage ranking. We report the official evaluation metrics for passage ranking, i.e., MRR@10 for Dev and NDCG@10 for DL19/DL20. For out-of-domain evaluation, we use 13 publicly available BEIR testsets~\cite{thakurbeir} covering different text domains. Again, we report the official evaluation metric NDCG@10.
All evaluations involve first constructing the index with the first-stage retriever, then retrieving passages with the retriever, followed by refining the ranklist using our trained rerankers.

\paragraph{Language Models Used.}
We conduct a comparative study between rerankers using state space models and several previously studied language models. Among encoder-only models, we use BERT~\cite{devlin-etal-2019-bert}, RoBERTa~\cite{liu2019roberta}, and ELECTRA~\cite{clarkelectra} with their \texttt{base} as well as \texttt{large} variants. For encoder-decoder models, we select both \texttt{base} and \texttt{large} variants of the BART model~\cite{lewis-etal-2020-bart}. Among decoder-only models, we compare with OPT~\cite{zhang2022opt}, and Llama3~\cite{dubey2024llama} models. The Llama3 model serves as an upper bound for transformer-based models, given that is the state-of-the-art pre-trained model at the 1B scale and high pre-training cost. We compare these with both Mamba-1 and Mamba-2-based rerankers at four different parameter settings. The details of the models used in our comparison study are shown in~\cref{tab:models}.

This extensive selection of pre-trained language models enables the comparison across different architecture types (e.g., encoder-only vs decoder-only),
% \vs{here and other places. Use "e.g.," instead of "ex:" to introduce examples}, 
pre-trained model scales (from 110M to 1.4B parameters), as well as different pre-training objectives (e.g., masked language modeling in BERT vs replaced token detection in ELECTRA). 
It is important to acknowledge that the pre-trained models are trained under different pre-training setups (such as varying datasets and hyperparameter configurations, etc.); comparing them is not entirely fair. Nevertheless, we can gain insights by evaluating different language models using a consistent fine-tuning approach.

\paragraph{Baselines.} 
We include MonoBERT~\cite{nogueira2019passage}, cross-SimLM~\cite{wang2022simlm}, MonoT5~\cite{nogueira2020document}, RankT5~\cite{zhuang2023rankt5} as well as the state-of-the-art RankLlama model~\cite{ma2023fine}.
\cref{appendix:additional_details} gives a detailed description of these methods .

\paragraph{Implementation Details.}
Our code is implemented in PyTorch~\citep{paszke2019pytorch} using the Huggingface library~\cite{wolf2019huggingface}. 
The weights of the pre-trained language models are obtained from the Huggingface Hub\footnote{\url{https://huggingface.co/models}}.
Wherever applicable, we use Flash Attention 2~\cite{dao2024flashattention}, gradient accumulation, and activation checkpointing. Note that the models are trained \emph{without} parameter-efficient fine-tuning techniques such as LoRA~\cite{hu2021lora} which is different from~\citet{ma2023fine}. 
We also do not investigate alternative compression techniques for improved parameter efficiency, such as low-rank factorization~\cite{gupta2024empirical}, and leave these avenues for future research.

We do not extensively tune hyperparameters; as discussed by prior works~\cite{boytsov2022understanding,ma2023fine} fine-tuning of reranking models is less sensitive to hyperparameters. 
We found the vanilla AdamW optimizer along with learning rate warm-up with linear scheduler to work for all training runs.
Refer to~\cref{appendixc:hyperparameters} for an overview of hyperparameters throughout the experiments.
Out implementation as well as checkpoints will be made public to facilitate reproducibility.

For the autoregressive models, including Mamba models, we provide input with the following template: \texttt{document}: \texttt{$\{d\}$} \texttt{;} \texttt{query}: \texttt{$\{q\}$} \texttt{;} \texttt{[EOS]}. The linear layer then takes the last layer representation of the \texttt{[EOS]} token and outputs the relevance score:
\begin{equation}
    f_{\theta}(q, d) = \mathrm{Linear} \big( \mathrm{model}(\mathrm{input})[-1]\big)
\end{equation}
For encoder-only and encoder-decoder models, we use a  different template: \texttt{[CLS]} \texttt{;} 
 \texttt{query}: \texttt{$\{q\}$} \texttt{;} \texttt{document}: \texttt{$\{d\}$}. The linear layer in this case uses the representation of the \texttt{[CLS]} token.

\begin{table*}[h!]
% \vspace{0pt}
\centering
\resizebox{0.99\textwidth}{!}{
\begin{tabular}{
lrlrrr
}
\toprule
\begin{tabular}[c]{@{}l@{}l} \textbf{Model} \\ \, \\ \end{tabular} &
\begin{tabular}[c]{@{}l@{}l} \textbf{Size} \\ \, \\ \end{tabular} & 
\begin{tabular}[c]{@{}l@{}l} \textbf{Retriever} \\ \, \end{tabular} & 
% \begin{tabular}[c]{@{}l@{}l} Top-$k$ \\ \, \end{tabular} & 
\begin{tabular}[c]{@{}l@{}l} \textbf{Dev} \\ \textbf{MRR@10} \\ \end{tabular} &
\begin{tabular}[c]{@{}l@{}l} \textbf{DL19} \\ \textbf{NDCG@10} \\ \end{tabular} &
\begin{tabular}[c]{@{}l@{}l} \textbf{DL20} \\ \textbf{NDCG@10} \\ \end{tabular} \\
\midrule
% BM25 & - & - & $|\mathcal{D}|$ & 18.4 & 50.6 & 48.0 \\
% bi-SimLM~\cite{wang2022simlm} & 110M & - & $|\mathcal{D}|$ & 39.1 & 69.8 & 69.2 \\
% GTR-base~\cite{ni-etal-2022-sentence} & 110M & - & $|\mathcal{D}|$ & 36.6 & - & - \\
% BGE-large-en-v1.5~\cite{xiao2023bge} & 335M & - & $|\mathcal{D}|$ & 35.7 & 70.8 & 70.7 \\
% OpenAI Ada2~\cite{neelakantan2022text} & \textbf{?} & - & $|\mathcal{D}|$ & 34.4 & 70.4 & 67.6 \\
% RepLlama~\cite{ma2023fine} & 7B & - & $|\mathcal{D}|$ & \textbf{41.2} & \textbf{74.3} & \textbf{72.1} \\ \hline

MonoBERT~\cite{nogueira2019passage} & 110 M & BM25 & 37.2 & 72.3 & 72.2 \\
cross-SimLM~\cite{wang2022simlm} & 110 M & bi-SimLM & 43.7 & 74.6 & 72.7 \\
MonoT5~\cite{nogueira2020document} & 220 M & BM25 & 38.1 & - & - \\
% MonoT5~\cite{nogueira2020document} & 3B & BM25 & 1000 & 39.8 & - & - \\
RankT5~\cite{zhuang2023rankt5} & 335 M & GTR & 42.2 & - & - \\
RankLlama~\cite{ma2023fine} & 7 B & RepLlama & \textbf{44.9}$\dag$ & \textbf{75.6}$\dag$ & \textbf{77.4}$\dag$ \\
% \rowcolor{lightgray}
\midrule
% \multicolumn{6}{l}{\emph{Results from this paper}} \\

BERT-base $^\textsc{E}$ & 110 M & \textsc{BGE} & 38.5 & 73.3 & 73.1 \\
RoBERTa-base $^\textsc{E}$ & 120 M & \textsc{BGE} & 39.1 & \textbf{75.4} & 72.0 \\
ELECTRA-base $^\textsc{E}$ & 105 M & \textsc{BGE} & \textbf{39.8} & 73.4 & \textbf{74.1} \\ 
BART-base $^\textsc{ED}$ & 130 M & \textsc{BGE} & 37.8 & 74.7 & 70.2 \\
OPT-125M $^\textsc{D}$ & 125 M & \textsc{BGE} & 35.2 & 70.6 & 69.2 \\
Mamba-1-130M $^\textsc{D}$ & 130 M & \textsc{BGE} & 37.8 & 73.7 & 70.5 \\
Mamba-2-130M $^\textsc{D}$ & 130 M & \textsc{BGE} & 37.0 & 73.8 & 70.8 \\ \midrule

BERT-large $^\textsc{E}$ & 330 M & \textsc{BGE} & 39.1 & \textbf{76.4} & 72.4 \\
RoBERTa-large $^\textsc{E}$ & 335 M & \textsc{BGE} & 37.8 & 75.1 & 69.4 \\
ELECTRA-large $^\textsc{E}$ & 320 M & \textsc{BGE} & 38.8 & 74.9 & 73.2 \\
% \rowcolor{lightgray}
% \multicolumn{7}{l}{\emph{\textbf{Encoder-decoder models}}} \\

BART-large $^\textsc{ED}$ & 385 M & \textsc{BGE} & \textbf{39.2} & 74.6 & 72.2 \\
% \rowcolor{lightgray}
% \multicolumn{7}{l}{\emph{\textbf{Decoder-only models}}} \\

OPT-350M $^\textsc{D}$ & 350 M & \textsc{BGE} & 36.3 & 72.1 & 68.9 \\
Mamba-1-370M $^\textsc{D}$ & 370 M & \textsc{BGE} & 38.9 & 74.7 & 72.5 \\
Mamba-2-370M $^\textsc{D}$ & 370 M & \textsc{BGE} & 38.6 & 75.8 & \textbf{74.0} \\ \midrule 
Mamba-1-790M $^\textsc{D}$ & 790 M & \textsc{BGE} & 38.2 & 76.4 & 72.9 \\
Mamba-2-780M $^\textsc{D}$ & 780 M & \textsc{BGE} & \textbf{39.0} & \textbf{76.8} & \textbf{73.6} \\ \midrule 
OPT-1.3B $^\textsc{D}$ & 1.3 B & \textsc{BGE} & 38.9 & 74.2 & 73.7 \\
Mamba-1-1.4B $^\textsc{D}$ & 1.4 B & \textsc{BGE} & 38.9 & 74.7 & 72.5 \\
Mamba-2-1.3B $^\textsc{D}$ & 1.3 B & \textsc{BGE} & 38.6 & 75.8 & 74.0 \\
Llama-3.2-1B $^\textsc{D}$ & 1.3 B & \textsc{BGE} & \textbf{40.4}$\ddag$ & \textbf{76.8}$\ddag$ & \textbf{76.2}$\ddag$ \\
\bottomrule
\end{tabular}
}
\caption{Results for passage reranking in-domain evaluation. We denote BGE-large-en-v1.5 as \textsc{BGE} for simplicity. We mark best results in each section bold; $\dag$ indicates the overall best result and $\ddag$ indicates the best result among our trained models. For the reranking threshold, RankLlama reranks top-100 results from RepLlama while other models reranks top-1000 results. Superscript $\textsc{E}$ denotes encoder-only model, \textsc{ED} denotes encoder-decoder model and \textsc{D} denotes decoder-only model.
}
\label{tab:results_passage}
% \vspace{-5pt}
\end{table*}
 
\subsection{Passage Reranking Results}
\label{subsec:passage_results}
\textbf{In-domain Evaluation.}
We show the in-domain passage reranking results in~\cref{tab:results_passage}. 
First, note that our trained models are comparable to previously reported results. For example, we report that BERT-base achieves 38.5, 73.3, 73.1 on Dev, DL19, DL20 respectively, compared to MonoBERT~\cite{nogueira2019passage}'s 37.2, 72.3 and 72.2. This suggests the correctness of our training setup. 

Between transformer models of different architectures, we notice that both encoder-only and encoder-decoder models outperform decoder-only models (OPT-125M and 350M in our case), despite OPT  being pre-trained with more tokens. We hypothesize the reason is that the bi-directional attention in encoders better capture the interaction between query and document tokens. But decoder-only models are easier to scale.

Between transformer and Mamba architectures, Mamba models are able to achieve strong performance. For example, despite being uni-directional, Mamba-2-370M achieves 38.6, 75.8, and 74.0 on three datasets compared to the best transformer-based model in that parameter range---BERT-large's 39.1, 76.4, and 72.4. The overall best transformer-based model---Llama-3.2-1B outperforms the Mamba models of similar size. However, note that Llama-3.2-1B is pre-trained on 15T tokens compared to Mamba model's 300B tokens. We conclude that Mamba models are competitive in the passage ranking task.

Among Mamba models, despite being trained on the same number of tokens, we notice overall that Mamba-2 achieves better performance than Mamba-1. A similar trend is shown in BEIR and document reranking results. In conclusion, Mamba-2 is a better SSM architecture compared to Mamba-1 for text reranking.

\begin{table*}[h!]
% \vspace{0pt}
\centering

\resizebox{\linewidth}{!}{
\begin{tabular}{@{}lrrrr|rrrr|rr@{}}
\toprule
& \multicolumn{1}{c}{BM25} & \multicolumn{1}{c}{MonoT5} & \multicolumn{1}{c}{RankT5} & \multicolumn{1}{c|}{RankLlama} & \multicolumn{1}{c}{ELECTRA} & \multicolumn{1}{c}{BART} & \multicolumn{1}{c}{Llama-3.2} & \multicolumn{1}{c|}{OPT} & \multicolumn{1}{c}{Mamba-1} & \multicolumn{1}{c}{Mamba-2} \\
\multicolumn{1}{l|}{Dataset} & - & 220M & 335M & 7B & 335M & 385M & 1.3B & 1.3B & 1.4B & 1.3B \\ \midrule
\multicolumn{1}{l|}{Arguana} & 39.7 & 19.4 & 22.3 & \textbf{56.0}$\dag$ & 14.6 & 18.0 & 32.7 & \textbf{35.7}$\ddag$ & 33.1 & \textbf{34.4} \\
\multicolumn{1}{l|}{Climate-FEVER} & 16.5 & 24.5 & 20.6 & \textbf{28.0}$\dag$ & 18.2 & 20.9 & 22.6 & \textbf{26.7}$\ddag$ & 22.6 & \textbf{26.2} \\
\multicolumn{1}{l|}{DBPedia} & 31.8 & 41.9 & 43.5 & \textbf{48.3}$\dag$ & 43.2 & 43.5 & 43.1 & \textbf{45.8}$\ddag$ & \textbf{45.8}$\ddag$ & \textbf{45.8}$\ddag$ \\
\multicolumn{1}{l|}{FEVER} & 65.1 & 80.1 & 83.5 & \textbf{83.9}$\dag$ & 76.8 & 77.5 & 72.9 & \textbf{83.0}$\ddag$ & 80.9 & \textbf{81.9} \\
\multicolumn{1}{l|}{FiQA} & 23.6 & 41.3 & 41.6 & \textbf{46.5}$\dag$ & 38.8 & 41.4 & 40.5  & \textbf{44.3}$\ddag$ & \textbf{43.3}$\ddag$ & \textbf{43.3}$\ddag$ \\
\multicolumn{1}{l|}{HotpotQA} & 63.3 & 69.5 & 71.3 & \textbf{75.3}$\dag$ & 68.6 & 71.9 & 69.2  & \textbf{74.9} & 75.8 & \textbf{76.3}$\ddag$ \\
\multicolumn{1}{l|}{NFCorpus} & 32.2 & \textbf{35.7} & 32.6 & 30.3 & 33.5 & 34.9 & \textbf{37.9} & 32.8 & 38.8 & \textbf{39.2}$\dag\ddag$ \\
\multicolumn{1}{l|}{NQ} & 30.6 & 56.7 & 59.6 & \textbf{66.3}$\dag$ & 49.2 & 51.0 & 48.2 & \textbf{52.6}$\ddag$ & 50.8 & \textbf{52.1} \\
\multicolumn{1}{l|}{Quora} & 78.9 & 82.3 & 82.2 & \textbf{85.0}$\dag$ & 79.3 & 73.6 & \textbf{84.9}$\ddag$ & 84.0 & 80.9 & \textbf{83.9} \\
\multicolumn{1}{l|}{SCIDOCS} & 14.9 & 16.4 & \textbf{18.2} & 17.8 & 16.5 & 17.0 & 17.7 & \textbf{17.8} & 19.0 & \textbf{19.6}$\dag\ddag$ \\
\multicolumn{1}{l|}{SciFact} & 67.9 & 73.5 & \textbf{74.9} & 73.2 & 65.9 & 65.7 & 71.7 & \textbf{72.7} & \textbf{77.4}$\dag\ddag$ & 76.8 \\
\multicolumn{1}{l|}{TREC-COVID} & 59.5 & 77.6 & 75.2 & \textbf{85.2}$\dag$ & 67.2 & 70.6 & 77.0 & \textbf{81.6} & \textbf{83.0}$\ddag$ & 79.9 \\
\multicolumn{1}{l|}{Touche-2020} & 44.2 & 27.7 & \textbf{45.9}$\dag$ & 40.1 & 34.3 & \textbf{34.9} & 32.8 & 33.2 & 36.7 & \textbf{37.7}$\ddag$ \\ \midrule
Average & 43.7 & 49.7 & 51.7 & \textbf{56.6}$\dag$ & 46.6 & 47.8 & 50.1 & \textbf{52.7} & 52.9 & \textbf{53.6}$\ddag$ \\ \bottomrule
\end{tabular}}
\label{tab:results_beir}
\caption{Results for passage reranking out-of-domain evaluation. We show results of the largest encoder-only, encoder-decoder model and decoder-only models. Full results are referred to~\cref{sec:full_beir_results}. We mark best results in each section bold; $\dag$ indicates the overall best result and $\ddag$ indicates best result among our trained models.
}
\end{table*}

\noindent
\textbf{Out-of-domain Evaluation.}
We report part of BEIR results in~\cref{tab:results_beir} and leave full results to~\cref{sec:full_beir_results}. Overall, Mamba models are able to achieve competitive performance compared to the transformer-based models of similar sizes. Specifically, Mamba-2-1.3B achieves 53.6 NDCG@10 averaged over 13 datasets compared to OPT-1.3B's 52.7. Compared to baselines, Mamba-based models are only outperformed by the much larger RankLlama\,---\,a 7B model based on 7B-sized retrieval model RepLlama. This reinforces our findings from  in-domain evaluation, suggesting the efficacy of Mamba models in the passage reranking task. 

One surprising observation is the underperformance of the Llama-3.2-1B model. This pre-trained model was not only trained on more tokens (15B) but was also trained on a much more diverse set of web documents. Ideally, a model pre-trained on a more diverse set of documents should perform better on out-of-domain evaluation sets, but we find that to not be the case with Llama-3.2-1B model.

\subsection{Document Reranking}
\label{subsec:doc_ranking}
Next, we discuss the experimental setup and results for the document reranking task. 
The setup closely aligns with that used for passage reranking, with specific differences highlighted where applicable.

\paragraph{Datasets and Evaluation Metrics.}
We use the document ranking subset from the MSMARCO dataset containing 320K training instances. We use Pyserini's implementation of BM25~\cite{robertson1995okapi}\footnote{\url{https://github.com/castorini/pyserini}} as the first-stage document retriever and use top-100 documents to uniformly sample 7 hard negatives for each positive query-document pair. 
We train two model variants\,---\,FirstP and LongP\,---\,which truncate the input at 512 and 1,536 tokens respectively.
Prior works~\cite{boytsov2022understanding,ma2023fine} note that longer training lengths only yield marginal performance improvements. So we do not experiment with them.

For evaluation, we use the official development set (\texttt{Dev}) containing  5,193 queries and report MRR@100 for comparison. For the TREC DL19/DL20~\cite{craswell2020overview,craswell2021overview} evaluation set that includes 43/45 queries, we use the official NDCG@10 as our evaluation metric.

\paragraph{Other Details.} 
Among the baselines, we include MonoT5~\cite{nogueira2020document}, RankT5~\cite{zhuang2023rankt5} along with RankLlama that is the current state-of-the-art model using 7 Billion parameters. We also report two baseline runs from~\citet{boytsov2022understanding}: BERT-base-FirstP and BERT-base-MaxP
% \vs{what is MaxP? The above text described FirstP and LongP} 
as a sanity check for our implementation. MaxP method first segments the long document into several shorter passages, then uses the maximum relevance of segmented passages as the relevance of the document.
We train document reranking models for each of the pre-trained models highlighted in~\cref{subsec:passage}. Note that as encoder-only models have a fixed context length (ex: BERT with 512), we do not have LongP variants for them.

\begin{table}[h!]
% \vspace{0pt}
\centering
\resizebox{\columnwidth}{!}{
\begin{tabular}{lrrrr}
\toprule
Model & Size & \multicolumn{1}{l}{Dev} & \multicolumn{1}{l}{DL19} & \multicolumn{1}{l}{DL20} \\
&  & \multicolumn{1}{l}{MRR@100} & \multicolumn{1}{l}{NDCG@10} & \multicolumn{1}{l}{NDCG@10} \\
\midrule
BERT-base-FirstP & 110 M & 39.4 & 63.1 & 59.8 \\
BERT-base-MaxP & 110 M & 39.2 & 64.8 & 61.5 \\
% Longformer-base-LongP~\cite{boytsov2022understanding} & 140M & BM25-Q2D  & 41.4 & 67.6 & 62.8 \\
MonoT5 & 3 B & 41.1 & - & - \\
RankLlama & 7 B & \textbf{50.3}$\dag$ & \textbf{67.7} & \textbf{67.4}$\dag$ \\
\rowcolor{lightgray}
\multicolumn{5}{l}{\emph{\textbf{FirstP models}}} \\
BERT-base $^\textsc{E}$ & 110 M &  \textbf{41.3} & 65.8 & 61.5 \\
RoBERTa-base $^\textsc{E}$ & 125 M &  39.4 & 65.5 & 59.3 \\
ELECTRA-base $^\textsc{E}$ & 105 M &  39.0 & 66.3 & 62.3 \\ 
BART-base $^\textsc{ED}$ & 130 M  &  37.5 & 63.9 & 59.9 \\
OPT-125M $^\textsc{D}$ & 125 M &  38.8 & 63.8 & 61.8 \\
Mamba-1-130M $^\textsc{D}$ & 130 M &  40.9 & 66.5 & \textbf{64.4} \\
Mamba-2-130M $^\textsc{D}$ & 130 M &  38.3 & \textbf{66.7} & 63.9 \\ \midrule
BERT-large $^\textsc{E}$ & 330 M &  40.1 & 65.9 & 61.4 \\
RoBERTa-large $^\textsc{E}$ & 355 M &  \textbf{43.3}$\ddag$ & 66.8 & 64.2 \\
ELECTRA-large $^\textsc{E}$ & 335 M &  40.3 & \textbf{67.8} & \textbf{64.9} \\
BART-large $^\textsc{ED}$ & 385 M &  40.3 & 64.7 & 61.6 \\
OPT-350M $^\textsc{D}$ & 350 M &  39.0 & 64.7 & 63.1 \\
Mamba-1-370M $^\textsc{D}$ & 370 M &  42.5 & 67.8 & 63.9 \\
Mamba-2-370M $^\textsc{D}$ & 370 M &  41.0 & 67.2 & 64.7 \\ \midrule 

Mamba1-790M $^\textsc{D}$ & 790 M &  \textbf{42.0} & 67.4 & \textbf{64.9} \\
Mamba2-780M $^\textsc{D}$ & 780 M &  \textbf{42.0} & \textbf{68.7} & 64.6 \\ \midrule
OPT-1B $^\textsc{D}$ & 1.3 B &  40.8 & 65.3 & 61.8 \\
Llama-3.2-1B $^\textsc{D}$ & 1.3 B &  40.6 & 67.6 & 60.8 \\ 
Mamba-1-1.3B $^\textsc{D}$ & 1.3 B &  OOM & OOM & OOM \\
Mamba-2-1.3B $^\textsc{D}$ & 1.3 B &  \textbf{42.1} & \textbf{68.3} & \textbf{64.6} \\

\rowcolor{lightgray}
\multicolumn{5}{l}{\emph{\textbf{LongP models}}} \\
OPT-125M $^\textsc{D}$ & 125 M &  38.8 & 63.8 & 61.8 \\
Mamba-1-130M $^\textsc{D}$ & 130 M &  \textbf{39.2} & 66.0 & 63.0 \\
Mamba-2-130M $^\textsc{D}$ & 130 M &  38.3 &\textbf{67.3}& \textbf{63.6} \\ \midrule
OPT-350M $^\textsc{D}$ & 350 M &  35.7 & 64.3 & 60.5 \\
Mamba-1-370M $^\textsc{D}$ & 370 M &  39.3 & \textbf{67.8} & 64.3 \\
Mamba-2-370M $^\textsc{D}$ & 370 M &  \textbf{41.4} & 67.3 & \textbf{65.1} \\ \midrule
Mamba-1-790M $^\textsc{D}$ & 790 M &  41.3 & 68.0 & 64.9 \\
Mamba-2-780M $^\textsc{D}$ & 780 M &  \textbf{42.2} & \textbf{70.0}$\dag\ddag$ & \textbf{66.9}$\ddag$ \\ \midrule
OPT-1.3B $^\textsc{D}$ & 1.3 B &  \textbf{41.8} & 68.0 & \textbf{63.9} \\
Llama-3.2-1B $^\textsc{D}$ & 1.3 B &  40.9 & \textbf{68.5} & 63.5 \\ 
Mamba-1-1.4B $^\textsc{D}$ & 1.4 B &  OOM & OOM & OOM \\
Mamba-2-1.3B $^\textsc{D}$ & 1.3 B &  OOM & OOM & OOM \\

\bottomrule
\end{tabular}
}
\caption{Results for document reranking.  We mark the best result in each section bold; $\dag$ marks the overall best result and $\ddag$ marks best result among our trained models. For the reranking threshold, MonoT5 reranks top-1000 documents from the retriever while others rerank top-100 results. Superscripts $\textsc{E}$, \textsc{ED} and \textsc{D} are the same as~\cref{tab:results_passage}. OOM denotes Out-Of-Memory Error.}
\label{tab:results_doc}
% \vspace{-5pt}
\end{table}

\subsection{Document Reranking Results}
\label{subsec:doc_results}
The task of document reranking necessitates using models that can process lengthy contexts. Although transformer models can accommodate long contexts through adjustments like improved positional embeddings~\cite{su2024roformer} or specialized training methods~\cite{xiong2024effective,dubey2024llama}, it remains unclear whether Mamba-based state space models possess this capability. Our experiments in document reranking aim to address this gap. Our document reranking results are shown in~\cref{tab:results_doc}.

We make two important observations. First, in terms of the task performance, Mamba-based rerankers are comparable to their Transformer-based counterparts for every parameter budget. Notably, among the sub-1 billion parameter models, the best model is the 780 million Mamba-2 model trained with 1536 context length. Second, while the Mamba-1 and Mamba-2 variants perform comparably for document reranking, we found that Mamba-2 models in general require less GPU memory. One such instance is the training run with 1.3B parameters for context length of 512 (FirstP settings)---Mamba-1 leads to OOM error but Mamba-2 does not. This echoes prior works' observation that Mamba-2 is more memory efficient during training compared to Mamba-1~\cite[\interalia]{dao2024transformers,yang2024parallelizing}.

\subsection{Training Throughput and Inference Speed}
\label{subsec:efficiency_results}

\begin{figure}[t!]
    \centering
    \includegraphics[width=\columnwidth]{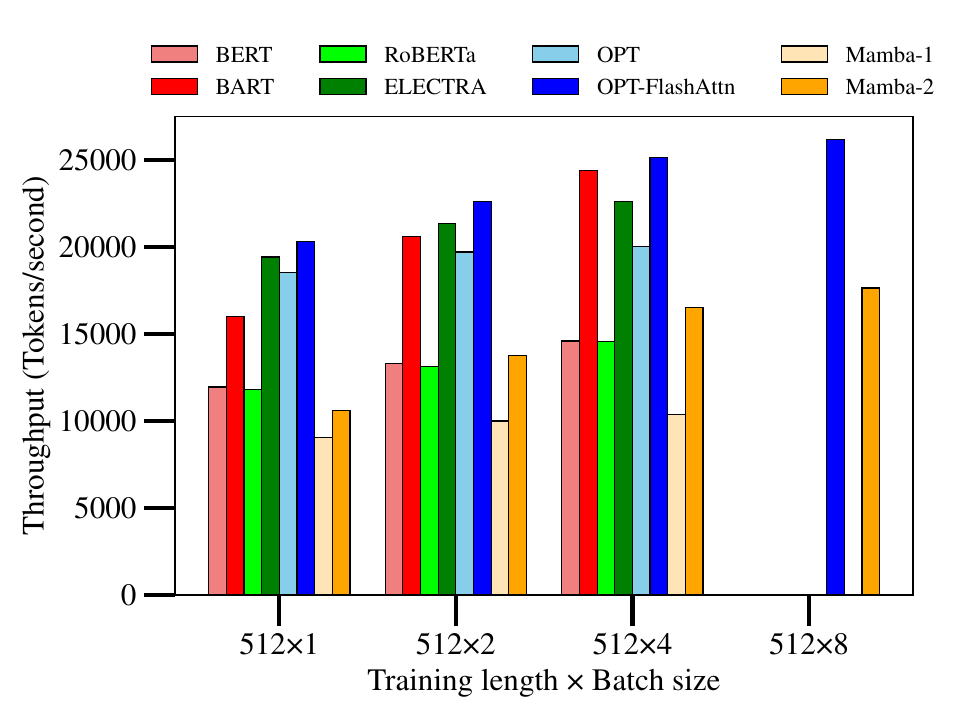}
    \caption{Training throughput comparison between models$\approx$330M. For batch\_size=8, all models except OPT-FlashAttn and Mamba-2 run out of memory with a 48 GB VRAM GPU.}
    \label{fig:throughput}
\end{figure}

To answer the \textbf{Efficiency RQ}, we evaluate the training throughput and inference speed of Mamba models and compare them to the transformer-based models. We perform this comparison with document ranking models as it involves a more challenging setting.
All the numbers reported here are measured on a server with Intel Xeon Gold 6230 CPU @2.1GHz and a single Nvidia A40 GPU (48 GB VRAM).

We measure training throughput (\#tokens/second) with 512 training length and the inference speed (\#queries/second) on MS MARCO document ranking dataset with max input length of 512 and 1,536. The results for the training throughput with different training batch sizes are shown in~\cref{fig:throughput}. The average inference speed over the queries from DL19 eval set is shown in~\cref{tab:inference_speed}.

First, observe that Mamba-2 has a much higher training throughput than Mamba-1. Additionally, since the Mamba-2 models are more memory efficient during training compared to Mamba-1, we do not notice an Out-Of-Memory (OOM) errors with Mamba-2\,---\,Mamba-1-370M does not train with batch size 8. 
The throughput of Mamba models is significantly worse than that of the transformer-based models. In other words, the Mamba-based models are much less efficient at training time.

\begin{table}[t!]
% \vspace{0pt}
\centering
\resizebox{\columnwidth}{!}{
\begin{tabular}{lrrr}
\toprule
\begin{tabular}[c]{@{}l@{}l} \textbf{Model} \\ \, \\ \end{tabular}
& \begin{tabular}[c]{@{}l@{}l} \textbf{Size} \\ \, \\ \end{tabular}
& \begin{tabular}[c]{@{}l@{}l} \textbf{Max.} \\ \textbf{Length} \\ \end{tabular}
& \begin{tabular}[c]{@{}l@{}l} \textbf{Queries. per} \\ \textbf{Second} ($\uparrow$) \\ \end{tabular} \\
\midrule 
BERT-large & 330 M & 512 & 0.65 \\
BART-large & 385 M & 512 & 0.65 \\
OPT-350M & 350 M & 512 & \textbf{0.69} \\
Mamba-1-370M & 370 M & 512 & 0.53 \\
Mamba-2-370M & 370 M & 512 & 0.56 \\
\midrule
OPT-350M & 370 M & 1536 & \textbf{0.45} \\
Mamba-1-370M & 370 M & 1536 & 0.40 \\
Mamba-2-370M & 370 M & 1536 & \textbf{0.45} \\
\midrule
OPT-1.3B & 1.3 B & 512 & 0.29 \\
Llama-3.2-1B & 1.3 B & 512 & \textbf{0.33} \\
Mamba-1-1.4B & 1.4 B & 512 & 0.25 \\
Mamba-2-1.3B & 1.3 B & 512 & 0.30 \\
\midrule
OPT-1.3B & 1.3 B & 1536 & 0.28 \\
Llama-3.2-1B & 1.3 B & 1536 & \textbf{0.31} \\
Mamba-1-1.4B & 1.4 B & 1536 & 0.24 \\
Mamba-2-1.3B & 1.3 B & 1536 & 0.29 \\

\bottomrule
\end{tabular}
}
\caption{Inference speed of different models. We use half precision and batch size 32 for all models.}
\label{tab:inference_speed}
\end{table}
As highlighted in prior research~\citep[\interalia]{waleffe2024empirical,gu2023mamba}, the true benefit of the Mamba models is realized with an improved inference speed. We do not observe this to be the case for the document reranking task (see~\cref{tab:inference_speed}). 
The main reason is that for inference, reranking only requires one single forward computation compared to multiple forward computations in autoregressive generation. We further discuss the deficiency of Mamba models in~\cref{subsec:profiling}.

\subsection{Profiling Inference Computation}
\label{subsec:profiling}
\begin{figure*}
    \begin{subfigure}{0.24\linewidth}
        \centering
        \includegraphics[width=\linewidth]{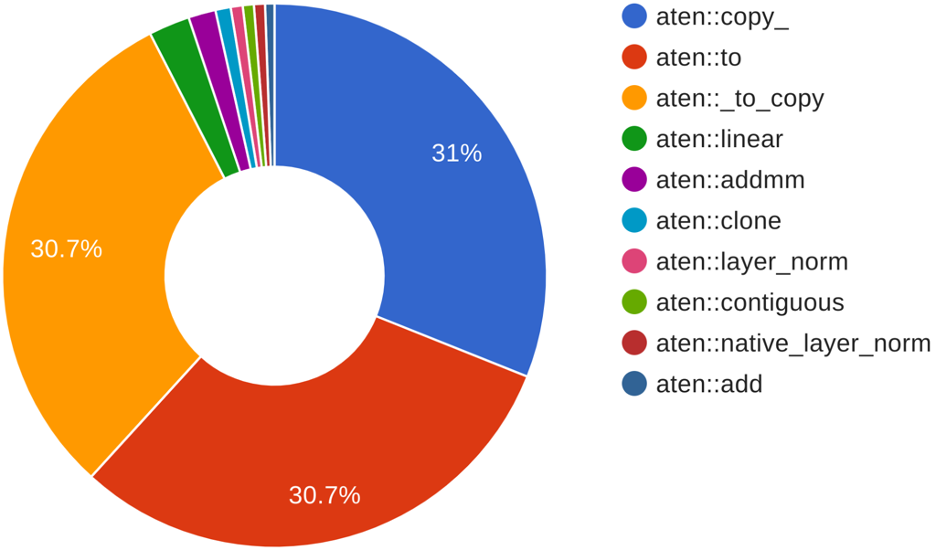}
        \caption{OPT-125M}
        \label{fig:profiling-opt-125m}
    \end{subfigure}
    \hfill
    \begin{subfigure}{0.24\linewidth}
        \centering
        \includegraphics[width=\linewidth]{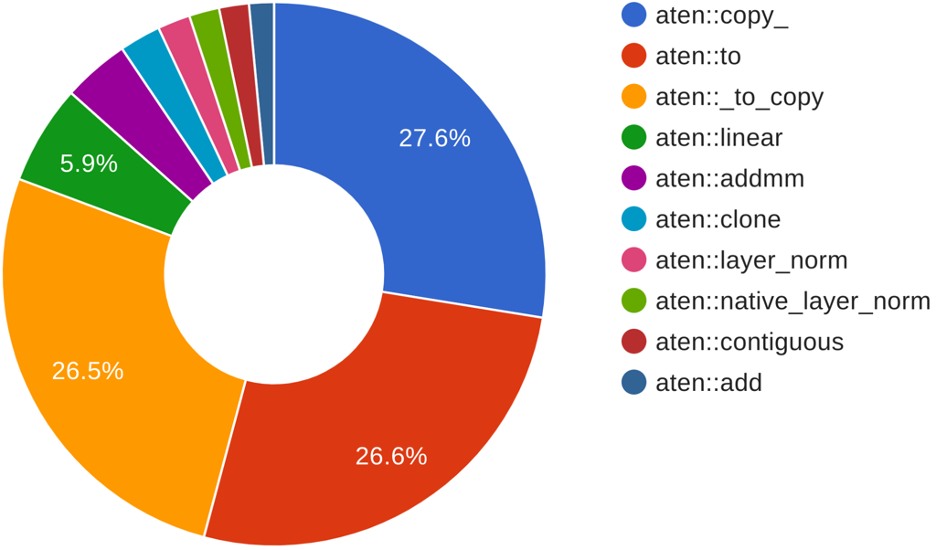}
        \caption{OPT-125M w/ FlashAttn}
        \label{fig:profiling-opt-125m-flashattn}
    \end{subfigure}
    \hfill
    \begin{subfigure}{0.24\linewidth}
        \centering
        \includegraphics[width=\linewidth]{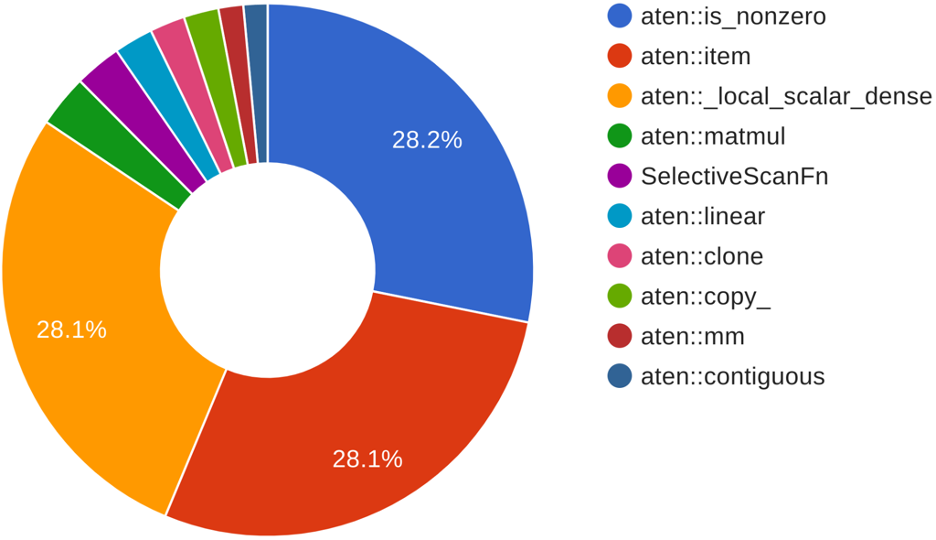}
        \caption{Mamba-1-130M}
        \label{fig:profiling-mamba1-130m}
    \end{subfigure}
    \hfill
    \begin{subfigure}{0.24\linewidth}
        \centering
        \includegraphics[width=\linewidth]{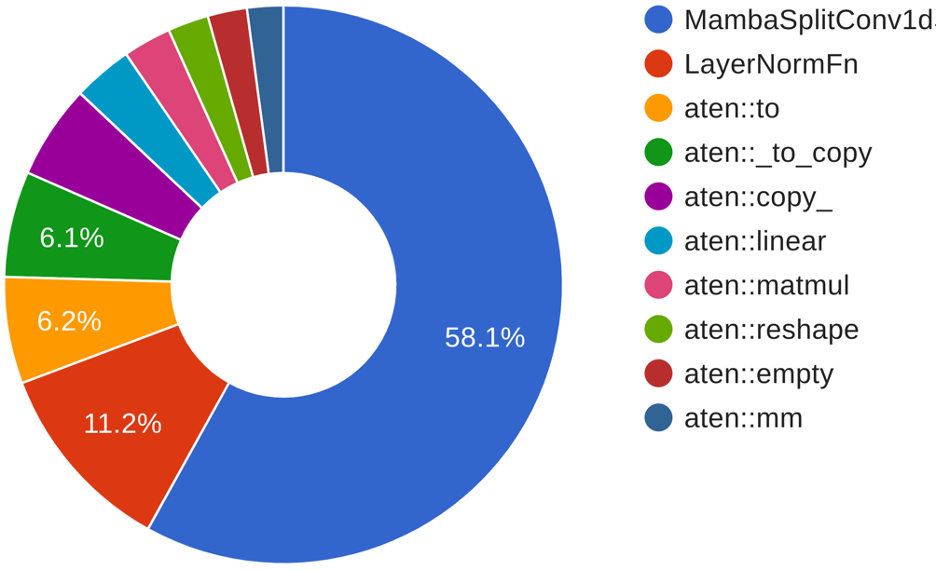}
        \caption{Mamba-2-130M}
        \label{fig:profiling-mamba2-130m}
    \end{subfigure}

    \begin{subfigure}{0.24\linewidth}
        \centering
        \includegraphics[width=\linewidth]{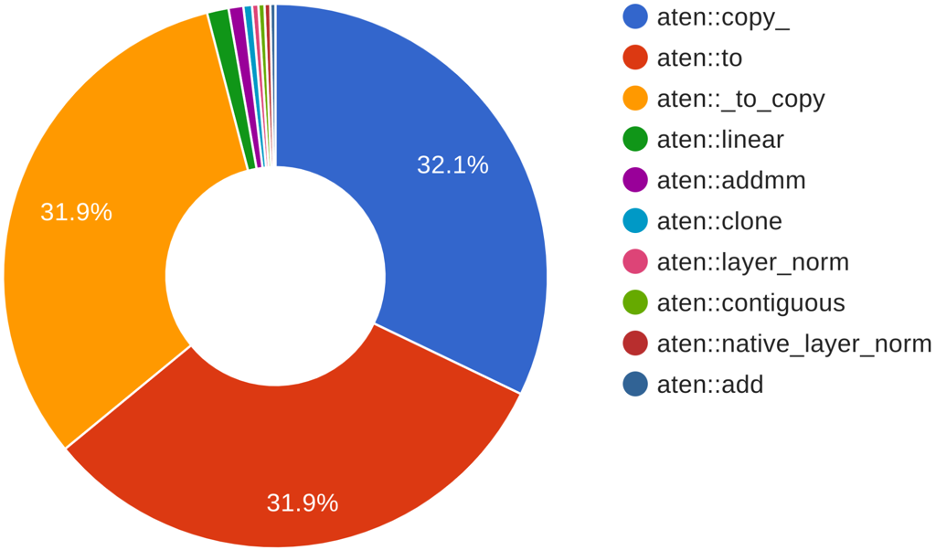}
        \caption{OPT-350M}
        \label{fig:profiling-opt-350m}
    \end{subfigure}
    \hfill
    \begin{subfigure}{0.24\linewidth}
        \centering
        \includegraphics[width=\linewidth]{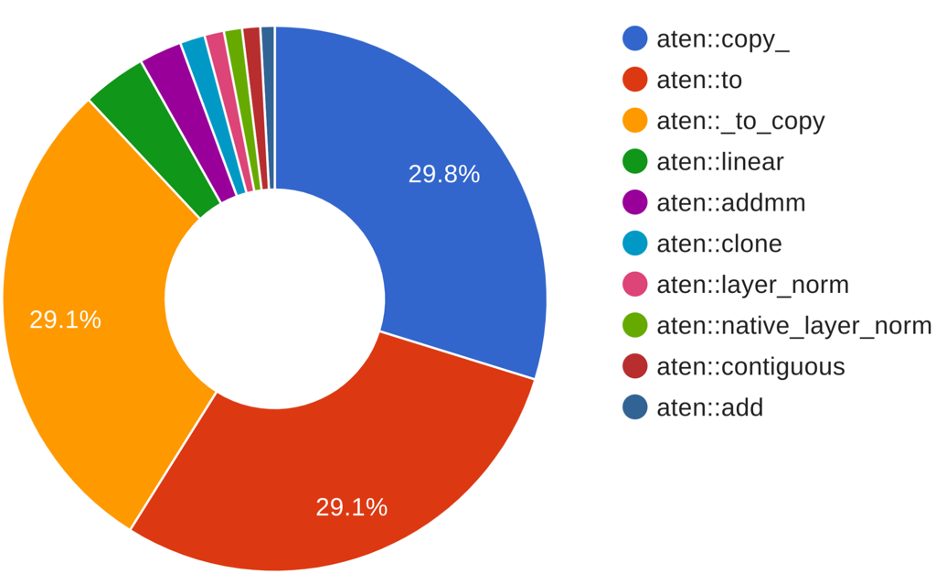}
        \caption{OPT-350M w/ FlashAttn}
        \label{fig:profiling-opt-350m-flashattn}
    \end{subfigure}
    \hfill
    \begin{subfigure}{0.24\linewidth}
        \centering
        \includegraphics[width=\linewidth]{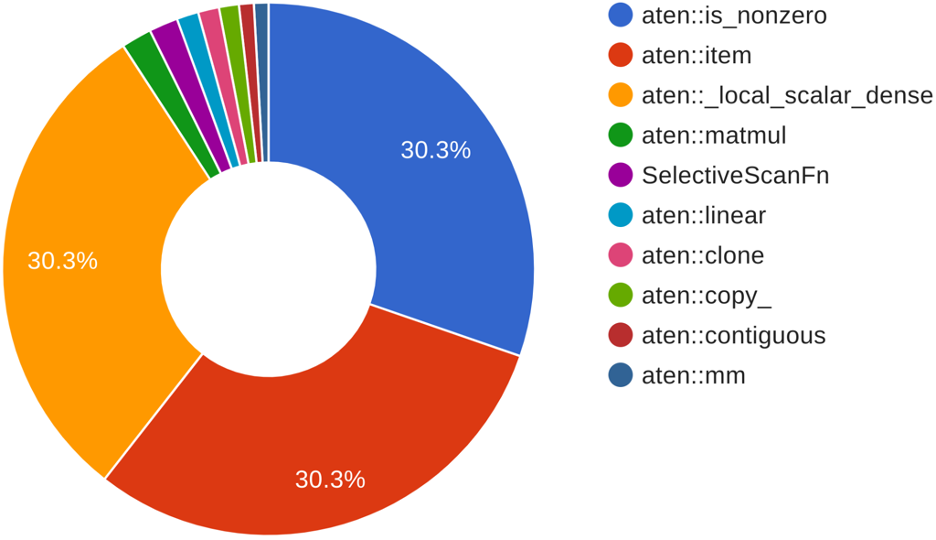}
        \caption{Mamba-1-370M}
        \label{fig:profiling-mamba1-370m}
    \end{subfigure}
    \hfill
    \begin{subfigure}{0.24\linewidth}
        \centering
        \includegraphics[width=\linewidth]{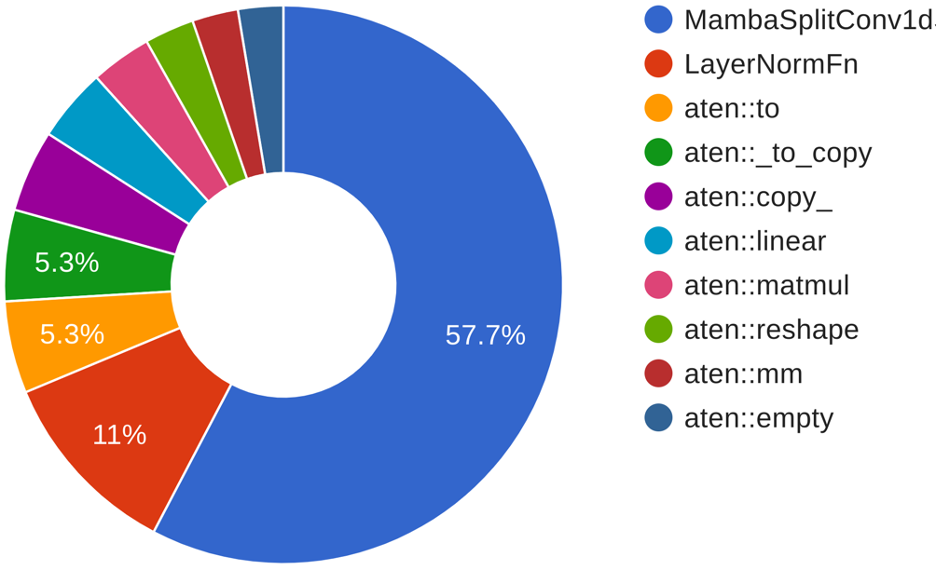}
        \caption{Mamba-2-370M}
        \label{fig:profiling-mamba2-370m}
    \end{subfigure}

\caption{
Inference profiling results for Mamba models versus OPT models of similar size.
}
\label{fig:profiling}
\vspace{-10pt}
\end{figure*}

To better understand the inference performance of Mamba models, we use the PyTorch profiler\footnote{https://pytorch.org/docs/stable/profiler.html} to analyze the execution time of Mamba models at the operator level, comparing it to Transformer-based models of similar sizes. As in~\cref{subsec:efficiency_results}, we use the DL19 document ranking evaluation set, an input length of 512, an evaluation batch size of 32, and the same hardware configuration. The results are presented in~\cref{fig:profiling}.

For Transformer-based models like OPT, I/O-related operators (e.g., \texttt{aten::copy\_}, \texttt{aten::to}, \texttt{aten::\_to\_copy}, etc.) account for the majority of the execution time. Flash Attention~\cite{dao2024flashattention} mitigates this by optimizing the I/O operations involved in attention computation, as seen in the reduced execution time for I/O-related operations in~\cref{fig:profiling-opt-125m-flashattn,fig:profiling-opt-350m-flashattn}. This optimization leads to a noticeable speedup in inference, highlighting the importance of improving I/O efficiency for Transformers.

In contrast, the total execution time of Mamba-1 is dominated by operators such as \texttt{aten::is\_nonzero}, \texttt{aten::item}, and \texttt{aten::\_local\_scalar\_dense}. The first operator, \texttt{aten::is\_nonzero}, checks whether tensors contain any non-zero elements, while \texttt{aten::item} and \texttt{aten::\_local\_scalar\_dense} are used to extract scalar values from tensors. This suggests that Mamba-1's architecture might suffer from computational inefficiencies due to an over-reliance on these scalar-extraction operations, which could be bottlenecking the performance. We hypothesize that these operations contribute to the model's overall computational deficiency, particularly in comparison to models that utilize more efficient tensor operations.
Mamba-2 improves upon this by parameterizing the Mamba-2 block, allowing for more effective utilization of matrix multiplication. This change is reflected in the elimination of the aforementioned scalar-extraction operators, with the new operator \texttt{MambaSplitConv1D} now accounting for over half of the total execution time. Mamba-2's shift towards matrix multiplication suggests a more balanced computational load, although it still doesn't fully close the gap in terms of inference speed compared to Transformer models with Flash Attention. This empirical evidence points to the need for further architectural refinement to optimize performance and better leverage compute-optimized hardware.

\section{Related Works}

\paragraph{Text Ranking with Pre-trained Transformers.} 
Fine-tuning pre-trained transformers has been the standard practice for text ranking tasks~\cite[\interalia]{yates2021pretrained,karpukhin-etal-2020-dense}. Combining the query and the document as the input, the model predicts a scalar score indicating the relevance.
Prior works have highlighted different aspects of training transformer-based text ranking models. \citet{nogueira2019passage,nogueira2020document,dai2019deeper} are among the first efforts to showcase the effectiveness of fine-tuning pre-trained transformer-based language models. 
\citet{gao2021rethink} studied the retrieval-reranking pipeline and recommended training rerankers by sampling negatives from the results of first-stage retrievers. 
\citet{li2023parade,hofstatter2021intra} studied the effectiveness of chunking and pooling in long document ranking with shorter context transformer models.
\citet{boytsov2022understanding} focused on benchmarking long context pre-trained transformers in long document ranking. Refer to~\cite{yates2021pretrained,xu2025survey} for detailed surveys. 

\paragraph{Transformer Alternatives.}
Different works have explored transformer alternative model architectures for sequence modeling. 
For example, S4~\cite{gu2021combining,smith2022simplified} demonstrate the effectiveness of structured state space models.
Recent works~\cite[\interalia]{peng-etal-2023-rwkv,yang2023gated,yang2024parallelizing,qin2024hgrn2} have vastly improved the computational bottleneck of RNN-alike architectures and have shown comparable performance to modern transformer architectures at a moderate scale of comparison. We refer readers to these works for more details. 

 Within the IR community, works have explored the possibility of using state space models as retriever~\cite{zhang2024mamba} and reranker~\cite{xu2024rankmamba}. Our study extends prior works with more comprehensive experiments and points out new directions. 
Additionally, there is extensive literature on domain generalization and robustness analysis of different types of transformer models in NLP~\citep{liang2023holistic,gupta2024whispers, awadalla2022exploring,gupta2023don}. Our domain generalization analysis with different types of transformer and state-space models augments this line of work for text ranking.

\section{Conclusion and Future Work}
This study investigates the suitability of Mamba architectures, a novel class of state space models, for text ranking. Our findings demonstrate that Mamba models, particularly Mamba-2, can achieve competitive performance compared to transformer-based models of comparable size, showcasing their potential as viable alternatives for sequence modeling in IR tasks. While Mamba architectures currently exhibit lower training and inference efficiency compared to transformers with flash attention, continuous advancements in model optimization and hardware acceleration have the potential to mitigate these limitations.

We picture two future directions of this work: the task direction and the model direction.
From the task perspective, the efficacy of state space models, including Mamba should be further examined in other IR tasks (e.g., text retrieval). 
From the model perspective, hybrid models~\cite{lieber2024jamba,team2024jamba,glorioso2024zamba,nemotronh2025} have shown promise in certain NLP tasks. We believe the effectiveness of hybrid models should be thoroughly tested. 
Additionally, optimization for state space models is an interesting challenge that may offer substantial improvements.

\section*{Limitations and Potential Risks}
This paper studies the efficacy of state space models in text ranking tasks. Our experiments are carried out by fine-tuning pre-trained language models, which differ in the pre-training corpus as well as pre-training FLOPs. Limited by hardware and budget, we are not able to carry out an apples-to-apples comparison with the exact same pre-training setup. We believe this leaves room for future direction.

This paper studies a well established task with publicly available datasets licensed for academic usage (see~\cref{appendix:licenses}). To the best of our knowledge this paper does not introduce potential risks.

\section*{Acknowledgements}
This material is based upon work supported in part by NSF under grants 2007398, 2217154, 2318550, 2205418, and 2134223.
This research is supported by the National Artificial Intelligence Research Resource (NAIRR) Pilot and the Delta advanced computing and data resource which is supported by the National Science Foundation (award NSF-OAC 2005572).
Ashim Gupta is supported by the Bloomberg Data Science Ph.D. Fellowship. 
Any opinions, findings, and conclusions or recommendations expressed in this material are those of the authors and do not necessarily reflect the views of the sponsers.

\bibliography{custom,anthology}

\appendix

\section{Dataset Artifacts and Licenses}
\label{appendix:licenses}
Four of the datasets we used in experiments (NFCorpus~\cite{boteva2016full}, FiQA-2018~\cite{maia201818}, Quora\footnote{\url{https://www.kaggle.com/c/quora-question-pairs}}, Climate-Fever~\cite{diggelmann2020climate}) do not report the dataset license in the paper or a repository.
For the rest of the datasets, we list their licenses below:
\begin{itemize}
    \item MS MARCO~\cite{bajaj2016ms}: MIT License for non-commercial research purposes.
    \item ArguAna~\cite{wachsmuth2018retrieval}: CC BY 4.0 license.
    \item DBPedia~\cite{hasibi2017dbpedia}: CC BY-SA 3.0 license.
    \item FEVER~\cite{thorne2018fever}: CC BY-SA 3.0 license.
    \item HotpotQA~\cite{yang2018hotpotqa}: CC BY-SA 4.0 license.
    \item NQ~\cite{kwiatkowski-etal-2019-natural}: CC BY-SA 3.0 license.
    \item SCIDOCS~\cite{cohan2020specter}: GNU General Public License v3.0 license.
    \item SciFact~\cite{wadden2020fact}: CC BY-NC 2.0 license.
    \item TREC-COVID~\cite{voorhees2021trec}: "Dataset License Agreement".
    \item Touche-2020~\cite{bondarenko2020overview}: CC BY 4.0 license.
\end{itemize}

\section{Additional Experiment Details}
\label{appendix:additional_details}

\subsection{Complexity Analysis of State Space Model}
\label{appendix:complexity}

We use the complexity analysis from~\cite{dao2024transformers}. For details, refer to Section 6 of~\citet{dao2024transformers}. Denote the sequence length as $L$ and state size as $N$, which means size $N$ per channel. We skip the \#channel dimension ($D$) for ease of comparison. SSD structure used in Mamba-2 is able to achieve better training and inference complexity, as reflected in our experiments (\cref{fig:throughput} and \cref{tab:inference_speed}).
\begin{table}[h!]
\vspace{0pt}
\centering
\resizebox{\columnwidth}{!}{
\begin{tabular}{
llll
}
\toprule
& Attention & SSM & SSD \\
\midrule
State size & $O(L)$ & $O(N)$ & $O(N)$ \\
Training FLOPs & $O(L^2N)$ & $O(LN^2)$ & $O(LN^2)$ \\
Inference FLOPs & $O(LN)$ & $O(N^2)$ & $O(N^2)$ \\ 
\midrule 
(Naive) memory & $O(L^2)$ & $O(LN^2)$ & $O(LN)$ \\
Matrix multiplication & \cmark & \xmark & \cmark \\
\bottomrule
\end{tabular}
}
\caption{Complexity analysis between state space structure and attention. 
}
\label{tab:complexity}
\vspace{-5pt}
\end{table}

\subsection{Baselines}
\subsubsection{Sparse and Dense Retrieval Methods}
For both document and passage retrieval, we include the classical BM25 baseline. For passage retrieval, bi-SimLM~\cite{wang2022simlm} is a competitive baseline that uses specialized pre-training with encoder-only transformer architecture for text retrieval task; GTR~\cite{ni-etal-2022-sentence} is based on T5~\cite{raffel2020exploring} architecture and is extensively fine-tuned for passage representations; BGE-large-en-v1.5~\cite{xiao2023bge} is based on BERT style encoder architecture and is fine-tuned with millions of synthetic query-passage pairs to achieve strong performance; OpenAI Ada2~\cite{neelakantan2022text} is a proprietary embedding model developed by OpenAI; RepLlama~\cite{ma2023fine} is based on Llama-2 language model~\cite{touvron2023llama} and is fine-tuned on the training split of MS MARCO datasets. It achieves state-of-the-art performance on passage retrieval.
For document retrieval, a common practice in literature is to segment long documents into several passages to fit into the 512 context length of BERT-style encoder-only transformer models. Each passage is scored individually and the relevance score of the document is an aggregation of individual passage's relevance scores. We include two such retrieval baselines: BM25-Q2D~\cite{nogueira2019document} uses the document expansion technique to enhance BM25's performance. CoCondenser-MaxP is based on CoCondenser technique~\cite{gao2022unsupervised} and uses max pooling for document relevance.

\subsubsection{Reranking Methods}
We include results from prior works as a comparison. For long document ranking, a common practice is to segment the long document into shorter passages and score them individually. For example,~\citet{dai2019deeper} referred to models only computing the relevance between query and the first document segment as FirstP, and methods that use the maximum relevance of passages within the document as the relevance of the document as MaxP. We refer to document ranking models that based on long context language models as LongP following~\citet{boytsov2022understanding}.

For document ranking, we include BERT-base-FirstP and BERT-base-MaxP from~\citet{boytsov2022understanding}. We also include another MaxP baseline MonoT5~\cite{pradeep2021expando} and a state-of-the-art LongP model RankLlama~\cite{ma2023fine}.

For passage ranking, we include results of MonoBERT~\cite{nogueira2019passage}, cross-SimLM~\cite{wang2022simlm}, MonoT5~\cite{nogueira2020document} and more recent RankT5~\cite{zhuang2023rankt5} and RankLlama~\cite{ma2023fine}. An additional note is these ranking models are coupled with different first-stage retrievers and with different training strategies. We refer to RankT5~\cite{zhuang2023rankt5} for a comprehensive study of loss functions and training strategies involved in training ranking models.

\section{Retrieval Results}
\label{sec:retrieval_results}
\begin{table*}[h!]
\vspace{0pt}
\centering

\resizebox{\textwidth}{!}{
\begin{tabular}{
lrrrrrrrr
}
\toprule
% \begin{tabular}[c]{@{}l@{}l} Model \\ \, \\ \end{tabular} &
% \begin{tabular}[c]{@{}l@{}l} Size \\ \, \\ \end{tabular} & 
% \begin{tabular}[c]{@{}l@{}l} Embed. Dim. \\ \, \\ \end{tabular} & 
% \begin{tabular}[c]{@{}l@{}l} Dev \\ MRR@10 \\ \end{tabular} &
% \begin{tabular}[c]{@{}l@{}l} DL19 \\ NDCG@10 \\ \end{tabular} &
% \begin{tabular}[c]{@{}l@{}l} DL20 \\ NDCG@10 \\ \end{tabular} \\

Model & Size & Embed. Dim. & \multicolumn{2}{c}{Dev} & \multicolumn{2}{c}{DL19} & \multicolumn{2}{c}{DL20} \\
& & & MRR@10 & Recall@1000 & NDCG@10 & Recall@1000 & NDCG@10 & Recall@1000 \\
\midrule
BM25 & - & - & 18.4 & 85.3 & 50.6 & 75.0 & 48.0 & 78.6 \\
bi-SimLM & 110M & 768 & 39.1 & 98.6 & 69.8 & - & 69.2 & - \\
GTR-base & 110M & 768 & 36.6 & 98.3 & - & - & - & - \\
GTR-XXL & 4.8B & 768 & 38.8 & 99.0 & - & - & - & - \\
BGE-large-en-v1.5 & 335M & 1024 & 35.7 & 97.6 & 70.8 & 84.5 & 70.7 & 83.0 \\
OpenAI Ada2 & \textbf{?} & 1536 & 34.4 & 98.6 & 70.4 & \textbf{86.3} & 67.6 & \textbf{87.1} \\
RepLlama & 7B & 4096 & \textbf{41.2} & \textbf{99.4} & \textbf{74.3} & - & \textbf{72.1} & - \\

\bottomrule
\end{tabular}
}
\caption{Passage retrieval performance of different retrieval models. We mark the best performance bold.}
\label{tab:results_passage_retrieval}
\vspace{-5pt}
\end{table*}
\begin{table*}[h!]
\vspace{0pt}
\centering

\resizebox{\textwidth}{!}{
\begin{tabular}{lrrrrrrrrr}
\toprule
Model & Size & Seg. Y/N & Embed. Dim.  & \multicolumn{2}{c}{Dev} & \multicolumn{2}{c}{DL19} & \multicolumn{2}{c}{DL20} \\
& & & & MRR@100 & Recall@1000 & NDCG@10 & Recall@100 & NDCG@10 & Recall@100 \\
\midrule
BM25 & - & N & - & 27.7 & 93.6 & 52.3 & 38.5 & 50.6 & 58.6 \\
BM25-Q2D & - & Y & - & 32.7 & 95.5 & 59.7 & \textbf{39.9} & 58.5 & \textbf{61.8} \\ 
CoCondenser-MaxP & 110M & Y & 768 & 42.5 & 93.9 & 64.8 & - & \textbf{64.0} & - \\ 
RepLlama & 7B & N & 4096 & \textbf{45.6} & \textbf{98.9} & \textbf{65.0} & - & 63.2 & - \\

\bottomrule
\end{tabular}
}
\caption{Document retrieval performance of different models.  We mark the best performance bold.}
\label{tab:results_doc_retrieval}
\vspace{-5pt}
\end{table*}
We show the passage retrieval results in~\cref{tab:results_passage_retrieval} and document retrieval results in~\cref{tab:results_doc_retrieval}.

\section{Hyperparameter Setting}
\label{appendixc:hyperparameters}
\begin{table*}[t]
\vspace{0pt}
\centering
\resizebox{\textwidth}{!}{
\begin{tabular}{
lrrrrrrrr
}
\toprule
Model & Size & Architecture & LR & Warmup & \#Epochs & Global BZ & AMP & FlashAttn \\
\rowcolor{lightgray}
\multicolumn{9}{l}{\emph{\textbf{Encoder-only Models} (Bi-directional)}} \\
BERT-base & 110M & Transformer & 2e-5 & 10\% & 2 & 8 & FP16 & \xmark \\
RoBERTa-base & 120M & Transformer & 2e-5 & 10\% & 2 & 8 & FP16 & \xmark \\
ELECTRA-base & 105M & Transformer & 2e-5 & 10\% & 2 & 8 & FP16 & \xmark \\ \hline
BERT-large & 330M & Transformer & 1e-5 & 10\% & 2 & 8 & FP16 & \xmark \\
RoBERTa-large & 335M & Transformer & 1e-5 & 10\% & 2 & 8 & FP16 & \xmark \\
ELECTRA-large & 320M & Transformer & 1e-5 & 10\% & 2 & 8 & FP16 & \xmark \\
\rowcolor{lightgray}
\multicolumn{9}{l}{\emph{\textbf{Encoder-Decoder Models} (Bi-directional)}} \\
BART-base & 130M & Transformer & 2e-5 & 10\% & 2 & 8 & FP16 & \xmark \\ \hline
BART-large & 385M & Transformer & 1e-5 & 10\% & 2 & 8 & FP16 & \xmark \\

\rowcolor{lightgray}
\multicolumn{9}{l}{\emph{\textbf{Decoder-only Models} (Uni-directional)}} \\
OPT-125M & 125M & Transformer & 2e-5 & 10\% & 2 & 8 & BF16 & \cmark \\
Mamba-1-130M & 130M & Mamba-1 & 2e-5 & 10\% & 2 & 8 & BF16 & \xmark \\
Mamba-2-130M & 130M & Mamba-2 & 2e-5 & 10\% & 2 & 4 & BF16 & \xmark \\ \hline
OPT-350M & 350M & Transformer & 1e-5 & 10\% & 2 & 8 & BF16 & \cmark \\
Mamba-1-370M & 370M & Mamba-1 & 1e-5 & 10\% & 2 & 4 & BF16 & \xmark \\
Mamba-2-370M & 370M & Mamba-2 & 1e-5 & 10\% & 2 & 4 & BF16 & \xmark \\ \hline
Mamba-1-790M & 790M & Mamba-1 & 1e-5 & 10\% & 1 & 4 & BF16 & \xmark \\
Mamba-2-780M & 780M & Mamba-2 & 1e-5 & 10\% & 1 & 4 & BF16 & \xmark \\ \hline
OPT-1.3B & 1.3B & Transformer & 1e-5 & 10\% & 1 & 4 & BF16 & \cmark \\
Mamba-1-1.4B & 1.4B & Mamba-1 & 1e-5 & 10\% & 1 & 4 & BF16 & \xmark \\
Mamba-2-1.3B & 1.3B & Mamba-2 & 1e-5 & 10\% & 1 & 4 & BF16 & \xmark \\
Llama-3.2-1B & 1.3B & Transformer++ & 1e-5 & 10\% & 1 & 4 & BF16 & \cmark \\

\bottomrule
\end{tabular}
}
\caption{Hyperparameters for passage reranking models. We use 10\% of the total training steps for linear learning rate warmup. Global BZ denotes global batch size; AMP denotes automatic mixed precision, FlashAttn denotes whether Flash Attention 2~\cite{dao2024flashattention} is used.
}
\label{tab:hyperparameters_passage}
\vspace{-5pt}
\end{table*}

\begin{table*}[t]
\vspace{0pt}
\centering
\resizebox{\textwidth}{!}{
\begin{tabular}{
lrrrrrrrr
}
\toprule
Model & Size & Architecture & LR & Warmup & \#Epochs & Global BZ & AMP & FlashAttn \\
\rowcolor{lightgray}
\multicolumn{9}{l}{\emph{\textbf{Encoder-only Models} (Bi-directional)}} \\
BERT-base & 110M & Transformer & 2e-5 & 10\% & 2 & 8 & FP16 & \xmark \\
RoBERTa-base & 120M & Transformer & 2e-5 & 10\% & 2 & 8 & FP16 & \xmark \\
ELECTRA-base & 105M & Transformer & 2e-5 & 10\% & 2 & 8 & FP16 & \xmark \\ \hline
BERT-large & 330M & Transformer & 1e-5 & 10\% & 2 & 8 & FP16 & \xmark \\
RoBERTa-large & 335M & Transformer & 1e-5 & 10\% & 2 & 8 & FP16 & \xmark \\
ELECTRA-large & 320M & Transformer & 1e-5 & 10\% & 2 & 8 & FP16 & \xmark \\
\rowcolor{lightgray}
\multicolumn{9}{l}{\emph{\textbf{Encoder-Decoder Models} (Bi-directional)}} \\
BART-base & 130M & Transformer & 2e-5 & 10\% & 2 & 8 & FP16 & \xmark \\ \hline
BART-large & 385M & Transformer & 1e-5 & 10\% & 2 & 8 & FP16 & \xmark \\

\rowcolor{lightgray}
\multicolumn{9}{l}{\emph{\textbf{Decoder-only Models} (Uni-directional)}} \\
OPT-125M & 125M & Transformer & 2e-5 & 10\% & 2 & 8 & BF16 & \cmark \\
Mamba-1-130M & 130M & Mamba-1 & 2e-5 & 10\% & 2 & 8 & BF16 & \xmark \\
Mamba-2-130M & 130M & Mamba-2 & 2e-5 & 10\% & 2 & 4 & BF16 & \xmark \\ \hline
OPT-350M & 350M & Transformer & 1e-5 & 10\% & 2 & 8 & BF16 & \cmark \\
Mamba-1-370M & 370M & Mamba-1 & 1e-5 & 10\% & 2 & 4 & BF16 & \xmark \\
Mamba-2-370M & 370M & Mamba-2 & 1e-5 & 10\% & 2 & 4 & BF16 & \xmark \\ \hline
Mamba-1-790M & 790M & Mamba-1 & 1e-5 & 10\% & 1 & 4 & BF16 & \xmark \\
Mamba-2-780M & 780M & Mamba-2 & 1e-5 & 10\% & 1 & 4 & BF16 & \xmark \\ \hline
OPT-1.3B & 1.3B & Transformer & 1e-5 & 10\% & 1 & 4 & BF16 & \cmark \\
Mamba-1-1.4B & 1.4B & Mamba-1 & 1e-5 & 10\% & 1 & 4 & BF16 & \xmark \\
Mamba-2-1.3B & 1.3B & Mamba-2 & 1e-5 & 10\% & 1 & 4 & BF16 & \xmark \\
Llama-3.2-1B & 1.3B & Transformer++ & 1e-5 & 10\% & 1 & 4 & BF16 & \cmark \\

\bottomrule
\end{tabular}
}
\caption{Hyperparameters for document reranking models. We use 10\% of the total training steps for linear learning rate warmup. Global BZ denotes global batch size; AMP denotes automatic mixed precision, FlashAttn denotes whether Flash Attention 2~\cite{dao2024flashattention} is used. Note for LongP models, we additionally use gradient accumulation and/or activation checkpoint techniques to maintain a reasonably large global batch size. Mamba-1-1.4B gets OOM in FirstP setting; Mamba-1-1.4B and Mamba-2-1.3B get OOM in LongP setting with batch size 1 despite all optimization techniques at our hands.
}
\label{tab:hyperparameters_doc}
\vspace{-5pt}
\end{table*}
We show the hyperparameters in~\cref{tab:hyperparameters_passage} and~\cref{tab:hyperparameters_doc}.

\section{Full BEIR Results}
\label{sec:full_beir_results}
We refer the full results on BEIR to~\cref{tab:full_beir_results}.
\begin{table*}[h!]
\vspace{0pt}
\centering

\resizebox{\linewidth}{!}{
\begin{tabular}{@{}lrrrr|rrrr@{}}
\toprule
& BM25 & MonoT5 & RankT5 & RankLlama & BERT-base & BART-base & RoBERTa-base & ELECTRA-base \\
\multicolumn{1}{l}{Dataset} & - & 220M & 335M & 7B & 110M & 130M & 120M & 105M \\ \midrule
\multicolumn{1}{l|}{Arguana} & 39.7 & 19.4 & 22.3 & 56.0 & 15.6 & 16.1 & 14.8 & 18.2 \\
\multicolumn{1}{l|}{ClimateFever} & 16.5 & 24.5 & 20.6 & 28.0 & 16.9 & 16.6 & 17.8 & 20.3 \\
\multicolumn{1}{l|}{DBPedia} & 31.8 & 41.9 & 43.5 & 48.3 & 38.5 & 42.5 & 42.1 & 42.1 \\
\multicolumn{1}{l|}{FEVER} & 65.1 & 80.1 & 83.5 & 83.9 & 73.9 & 72.9 & 70.9 & 78.2 \\
\multicolumn{1}{l|}{FiQA} & 23.6 & 41.3 & 41.6 & 46.5 & 34.6 & 38.4 & 36.4 & 40.1 \\
\multicolumn{1}{l|}{HotpotQA} & 63.3 & 69.5 & 71.3 & 75.3 & 66.0 & 69.7 & 70.8 & 68.9 \\
\multicolumn{1}{l|}{NFCorpus} & 32.2 & 35.7 & 32.6 & 30.3 & 29.3 & 32.7 & 26.1 & 29.9 \\
\multicolumn{1}{l|}{NQ} & 30.6 & 56.7 & 59.6 & 66.3 & 45.2 & 48.6 & 49.6 & 50.1 \\
\multicolumn{1}{l|}{Quora} & 78.9 & 82.3 & 82.2 & 85.0 & 75.8 & 75.3 & 74.8 & 79.3 \\
\multicolumn{1}{l|}{SCIDOCS} & 14.9 & 16.4 & 18.2 & 17.8 & 16.1 & 15.8 & 15.4 & 17.1 \\
\multicolumn{1}{l|}{SciFact} & 67.9 & 73.5 & 74.9 & 73.2 & 65.3 & 67.7 & 61.3 & 66.3 \\
\multicolumn{1}{l|}{TREC-COVID} & 59.5 & 77.6 & 75.2 & 85.2 & 67.8 & 70.3 & 70.9 & 72.3 \\
\multicolumn{1}{l|}{Touche-2020} & 44.2 & 27.7 & 45.9 & 40.1 & 30.7 & 33.2 & 30.1 & 33.3 \\ \midrule
\multicolumn{1}{l|}{Average} & 43.7 & 49.7 & 51.7 & 56.6 & 44.3 & 46.1 & 44.7 & 47.4 \\ \bottomrule
\end{tabular}
}
\bigskip

\resizebox{\linewidth}{!}{

\begin{tabular}{@{}lrrrrrrrr@{}}
\toprule
& OPT-125M & Mamba-1-130M & Mamba-2-130M & BERT-large & BART-large & RoBERTa-large & ELECTRA-large & OPT-350M \\
\multicolumn{1}{l}{Dataset} & 125M & 130M & 130M & 330M & 385M & 335M & 320M & 350M \\ \midrule
\multicolumn{1}{l|}{Arguana} & 10.1 & 32.8 & 33.8 & 19.5 & 18.0 & 15.4 & 14.6 & 21.0 \\
\multicolumn{1}{l|}{ClimateFever} & 5.9 & 21.0 & 23.1 & 23.4 & 20.9 & 15.1 & 18.2 & 8.1 \\
\multicolumn{1}{l|}{DBPedia} & 17.6 & 43.8 & 43.7 & 43.1 & 43.5 & 42.7 & 43.2 & 23.0 \\
\multicolumn{1}{l|}{FEVER} & 9.5 & 76.6 & 76.3 & 79.5 & 77.5 & 71.9 & 76.8 & 19.8 \\
\multicolumn{1}{l|}{FiQA} & 11.2 & 38.9 & 40.7 & 38.2 & 41.4 & 36.4 & 38.8 & 16.1 \\
\multicolumn{1}{l|}{HotpotQA} & 31.7 & 72.2 & 72.8 & 70.2 & 71.9 & 66.8 & 68.6 & 48.1 \\
\multicolumn{1}{l|}{NFCorpus} & 10.2 & 36.3 & 37.2 & 35.0 & 34.9 & 27.7 & 33.5 & 12.9 \\
\multicolumn{1}{l|}{NQ} & 22.1 & 48.3 & 48.3 & 51.5 & 51.0 & 48.2 & 49.2 & 29.0 \\
\multicolumn{1}{l|}{Quora} & 34.5 & 85.1 & 84.5 & 76.6 & 73.6 & 82.1 & 79.3 & 60.2 \\
\multicolumn{1}{l|}{SCIDOCS} & 5.2 & 17.4 & 17.4 & 16.8 & 17.0 & 15.5 & 16.5 & 7.9 \\
\multicolumn{1}{l|}{SciFact} & 9.7 & 72.2 & 73.0 & 68.8 & 65.7 & 55.4 & 65.9 & 28.6 \\
\multicolumn{1}{l|}{TREC-COVID} & 51.9 & 75.9 & 79.0 & 68.0 & 70.6 & 70.8 & 67.2 & 57.3 \\
\multicolumn{1}{l|}{Touche-2020} & 10.4 & 36.4 & 36.3 & 48.6 & 34.9 & 29.6 & 34.3 & 16.1 \\ \midrule
\multicolumn{1}{l|}{Average} & 17.7 & 50.5 & 51.2 & 49.2 & 47.8 & 44.4 & 46.6 & 26.8 \\ \bottomrule
\end{tabular}
}

\bigskip

\resizebox{\linewidth}{!}{
\begin{tabular}{@{}lrrrrrrrr@{}}
\toprule
& Mamba-1-370M & Mamba-2-370M & Mamba-1-790M & Mamba-2-780M & OPT-1.3B & Llama-3.2-1B & Mamba-1-1.4B & Mamba-2-1.3B \\
\multicolumn{1}{l}{Dataset} & 370M & 370M & 790M & 780M & 1.3B & 1.3B & 1.4B & 1.3B \\ \midrule
\multicolumn{1}{l|}{Arguana} & 33.3 & 34.8 &34.4& 33.7 & 35.7 & 32.7 & 33.1 & 34.4 \\
\multicolumn{1}{l|}{ClimateFever} & 23.3 & 25.4 &24.7& 23.9 & 26.7 & 22.6 & 22.6 & 26.2 \\
\multicolumn{1}{l|}{DBPedia} & 45.8 & 46.0 &46.1& 46.4 & 45.8 & 43.1 & 45.8 & 45.8 \\
\multicolumn{1}{l|}{FEVER} & 76.5 & 79.1 &81.8& 80.4 & 83.0 & 72.9 & 80.9 & 81.9 \\
\multicolumn{1}{l|}{FiQA} & 42.4 & 41.5 &44.8& 43.6 & 44.3 & 40.5 & 43.3 & 43.3 \\
\multicolumn{1}{l|}{HotpotQA} & 75.7 & 75.0 &75.6& 76.2 & 74.9 & 69.2 & 75.8 & 76.3 \\
\multicolumn{1}{l|}{NFCorpus} & 37.9 & 39.1 &41.0& 39.9 & 32.8 & 37.9 & 38.8 & 39.2 \\
\multicolumn{1}{l|}{NQ} & 51.0 & 51.9 &53.4& 52.8 & 52.6 & 48.2 & 50.8 & 52.1 \\
\multicolumn{1}{l|}{Quora} & 86.0 & 83.5 &86.0& 84.4 & 84.0 & 84.9 & 80.9 & 83.9 \\
\multicolumn{1}{l|}{SCIDOCS} & 18.6 & 19.1 &19.1& 19.5 & 17.8 & 17.7 & 19.0 & 19.6 \\
\multicolumn{1}{l|}{SciFact} & 75.2 & 76.0 &77.7& 77.1 & 72.7 & 71.7 & 77.4 & 76.8 \\
\multicolumn{1}{l|}{TREC-COVID} & 82.7 & 81.2 &82.7& 85.1 & 81.6 & 77.0 & 83.0 & 79.9 \\
\multicolumn{1}{l|}{Touche-2020} & 48.6 & 36.1 &39.6& 37.5 & 33.2 & 32.8 & 36.7 & 37.7 \\ \midrule
\multicolumn{1}{l|}{Average} & 53.6 & 53.0 &54.4& 53.9 & 52.7 & 50.1 & 52.9 & 53.6 \\ \bottomrule
\end{tabular}
}
\caption{Full results for passage ranking out-of-domain evaluation. 
}
\label{tab:full_beir_results}
\end{table*}

\end{document}